%% file: ECCV_submit.tex
\documentclass[runningheads]{llncs}

\usepackage[final]{eccv}

\usepackage{eccvabbrv}

\usepackage{graphicx}
\usepackage{booktabs}

\usepackage[accsupp]{axessibility}  
\usepackage{url}
\usepackage{makecell}
\usepackage{float} 
\usepackage{amsmath}
\usepackage[dvipsnames]{xcolor}
\definecolor{col1}{RGB}{232, 161, 148}
\definecolor{col2}{RGB}{148, 187, 232}
\usepackage{colortbl}   
\usepackage{bm}
\usepackage{mathrsfs} 
\usepackage{dsfont}
\usepackage{multirow}
\usepackage{graphicx}
\usepackage[numbers]{natbib}
\usepackage{caption}
\usepackage{amsfonts}
\usepackage{xcolor}
\usepackage{booktabs}
\usepackage{tikz}
\usepackage{xfp}
\usepackage{wrapfig}
\usepackage{indentfirst}
\usepackage{capt-of}
\usepackage{annotate_equations}

 \renewcommand{\eqnannotationtext}[1]{      
    \sffamily\fontsize{6pt}{7pt}\selectfont #1\strut}

\usepackage{hyperref}

\usepackage{orcidlink}
\newcommand{\ourmethod}{\textsc{GEM-4D}}

\begin{document}

\title{GEM-4D: Geometry-Enhanced Video World Models for Robot Manipulation} 

\titlerunning{GEM-4D}






\author{%
  Kaichen Zhou$^{1,2,*}$ \ \ Yuzhen Chen$^{1,*}$ \ \ Fangneng Zhan$^{2}$ \\ 
  Hang Hua$^{4}$ \ \ Grace Chen$^{1}$ \ \ Xinhai Chang$^{2}$ \ \ Ao Qu$^{2}$ \\ 
  Yilun Du$^{1}$ \ \  Zhuang Liu$^{3}$ \ \ Paul Pu Liang$^{2,\dagger}$ \ \ Mengyu Wang$^{1,\dagger}$ \\[0.8em]
  $^{1}$Harvard University \\
  $^{2}$Media Lab and EECS, MIT \\
  $^{3}$Princeton University \\
  $^{4}$MIT-IBM Watson AI Lab \\
}
\authorrunning{K. Zhou et al.}
\maketitle

\renewcommand{\thefootnote}{}
\footnotetext{$^*$Equal contribution as first authors. $^\dagger$Joint supervision.}

\begin{center}
\vspace{-1.1cm}
\includegraphics[width=\linewidth]{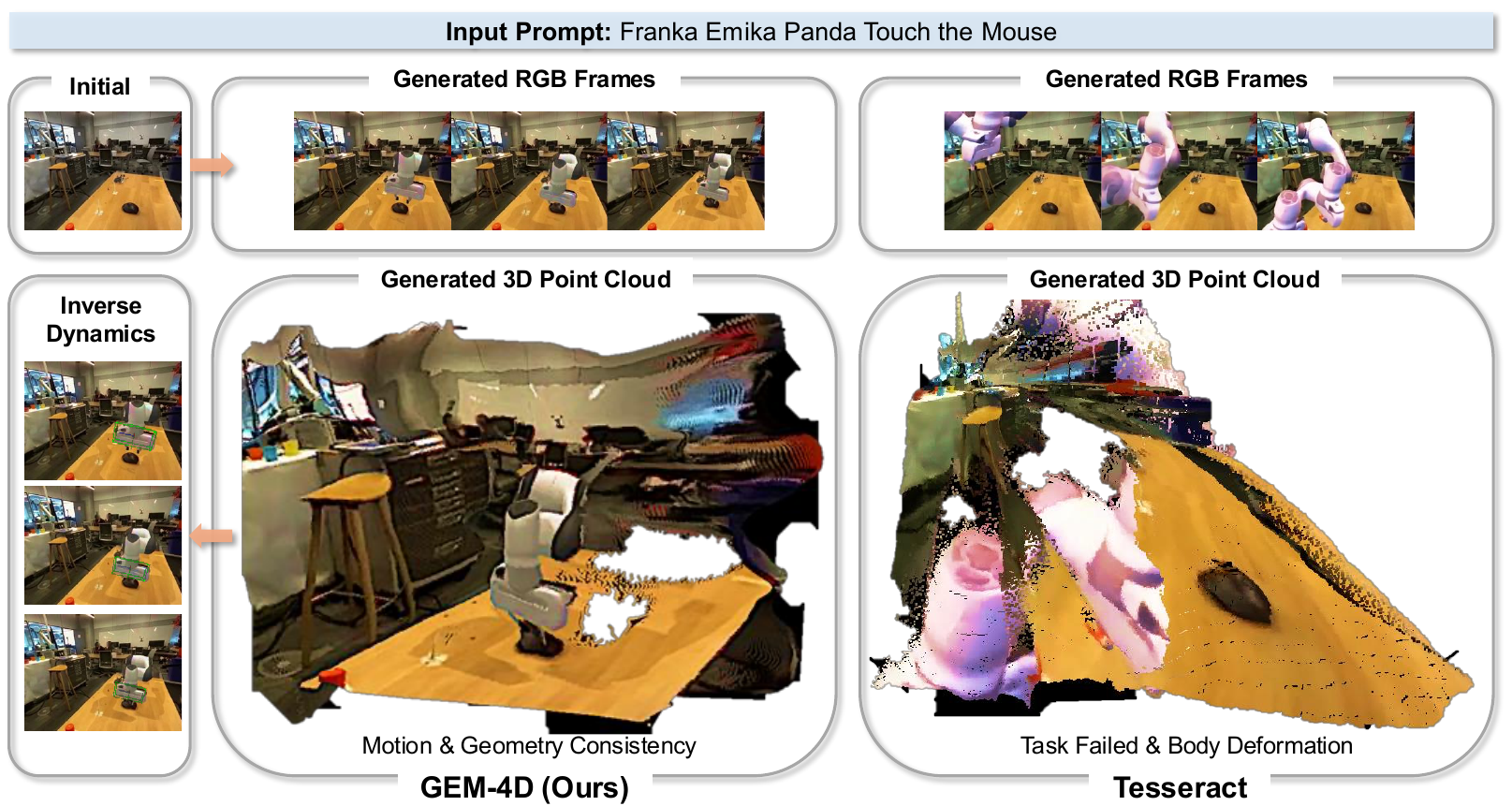}
\end{center}
\vspace{-0.5cm}
\captionsetup{type=figure}
\captionof{figure}{%
  Given an instruction and an initial observation, our model predicts future frames while preserving geometric consistency. Compared with the Tesseract~\cite{zhen2025tesseract} baseline (right), our approach produces more structurally coherent scene evolution.
  \label{fig:tesser}
}
\vspace{0.3cm}

\begin{abstract}
Video world models can generate realistic futures from a single instruction, but they often fail to track the same physical points consistently across time. 
As a result, the generated videos appear plausible, yet lack the physical grounding required for reliable action execution, such as robot manipulation.
We present \textsc{GEM-4D}, a geometry-grounded video world model that resolves this limitation by injecting dense 4D correspondence supervision distilled from a pretrained geometry foundation model into the video generative backbone during training.
This supervision enables the video world model to jointly capture appearance and geometric structure while retaining a single-stream architecture with no additional inference cost.
We further introduce an inverse dynamics module that converts correspondence-consistent video rollouts into executable robot trajectories, enabling direct deployment in both real-world and simulated manipulation.
\textsc{GEM-4D} achieves state-of-the-art performance on both video prediction and geometric consistency across both simulation and realistic scenarios and improves real-world manipulation success from 61\% to 81\%.
Additional results are available at the \url{https://gem-4d.github.io/}. 
\end{abstract}

\setlength{\parindent}{15pt} 

\input{section/introduction} 
\input{section/related}
\input{section/method}
\input{section/exp}
\input{section/conclusion}

%
%

\bibliographystyle{splncs04}
\bibliography{egbib}


\end{document}

%% file: section/introduction.tex
\section{Introduction}
\noindent General-purpose robotic manipulation requires policies that can generalize across diverse scenes, objects, tasks, and even embodiments. 
Vision-language-action policies~\cite{kim2024openvla, black2024pi0} have shown strong progress by directly mapping observations and language instructions to actions, but such generalization often depends on large-scale robot demonstration data and embodiment-specific training or fine-tuning. 
Video world models~\cite{zhen2025tesseract, guo2025ctrl,du2023learning, du2023video, zhou2024robodreamer,bharadhwaj2024gen2act, wu2024ivideogpt, zhu2025aether} offer a complementary path.
By predicting task-conditioned future observations from the current scene and instruction, they can provide a more general visual representation of how an interaction may unfold, potentially reducing reliance on task- and embodiment-specific action supervision. 


However, generic video generation models~\cite{wan2025wan,yang2024cogvideox} primarily emphasize visual realism, whereas using a video world model for robot planning requires more than photorealistic prediction. 
To derive executable actions from predicted futures, a robot must recover precise object and end-effector motions~\cite{bharadhwaj2024track2act}. 
This requires the generated video to preserve reliable inter-frame correspondences, such that pixels corresponding to the same 3D surface point follow physically consistent trajectories across time~\cite{huang2026pointworld}.

Current video diffusion models~\cite{peebles2023scalable, blattmann2023stable}, trained largely with pixel- or latent-space reconstruction objectives, provide no explicit guarantee of such consistency~\cite{wu2025geometry,liu2024sora}. 
They can produce photorealistic videos in which rigid objects deform non-rigidly, contacts drift, and depth varies inconsistently across frames~\cite{zhang2025videorepa}---errors that may be visually subtle but can fundamentally break action extraction. 
This limitation is structural: reliable inter-frame correspondence depends on camera motion, scene depth, and object motion, whereas standard pixel- or latent-space generation losses do not explicitly constrain these factors.
Existing remedies address this limitation only partially. 
Explicit 4D-supervised methods~\citep{zhen2025tesseract,zhu2025aether} add predictions such as RGB, depth, and surface normals, thereby constraining certain geometric quantities. However, they require large-scale annotations and still do not provide a unified correspondence signal that jointly accounts for camera motion, scene depth, and object motion.

Our key observation is that the geometric factors governing inter-frame correspondence are already encoded by modern 4D geometry foundation models. 
Models such as PAGE-4D~\cite{zhou2026page}, Depth Anything V3~\cite{lin2025depth}, VGGT~\cite{wang2025vggt} and VGGT-omega~\cite{wang2026vggt} estimate dense geometry and camera motion from video, so their intermediate representations capture depth, viewpoint change, and object motion in a unified form. 
Rather than supervising correspondence explicitly, we distill these geometry-aware representations into the video backbone~\cite{yu2025repa}. 
This feature-level supervision encourages the generative model to encode the camera, depth, and motion structure needed for consistent inter-frame correspondence, yielding predicted futures that are more reliable for robot action extraction.


We present \textsc{GEM-4D}, a world model that operationalizes this principle via feature-level distillation.
During training, a video diffusion transformer is paired with a geometry branch that predicts representations from a pretrained 4D geometry foundation model, conditioned on intermediate video features.
This forces the backbone to internalize correspondence-consistent geometric structure without modifying its output space or adding parameters.
The coupling is asymmetric and efficient: the geometry branch reads from video features but never writes back, and is discarded entirely at inference, yielding a single-stream generator with zero additional cost.
We further introduce an inverse dynamics module that converts correspondence-consistent rollouts into executable 6-DoF end-effector trajectories using off-the-shelf vision foundation models, closing the loop from language instruction to real-world manipulation without task-specific training.

\textsc{GEM-4D} achieves state-of-the-art performance on both video prediction and geometric consistency across both simulation and realistic scenarios and improves real-world manipulation success from 61\% to 81\%.
Our contributions are summarized as follows:
\begin{enumerate}
    \item \textbf{Principle.}
    We formalize the connection between geometry foundation model representations and inter-frame correspondences, showing that geometry supervision acts as a representation-level regularizer that encourages the video backbone to encode correspondence-consistent structure.

    \item \textbf{Architecture.}
    We introduce \textsc{GEM-4D}, a dual flow-matching framework that distills 4D geometry features into a video backbone via asymmetric conditioning, thereby achieving correspondence-aware generation at zero additional inference cost.

    \item \textbf{System.}
    We close the loop from world model to robot control: an inverse dynamics module extracts executable trajectories from generated rollouts, achieving 81\% success on real-world Droid tasks (+20 points over the strongest baseline) and 63–82\% success on RLBench.
\end{enumerate}

%% file: section/related.tex

\section{Related Work}

\subsection{Video World Models for Embodied Control}

\noindent Learning world models that predict future observations has become a central paradigm for embodied decision-making~\citep{du2023learning, yang2023learning, du2023video, zhou2024robodreamer, bharadhwaj2024gen2act, chi2024eva, qi2025strengthening,chen2025large,fu2025learning,qian2025wristworld,huang2026skill,wu2026multiworld, wu2024ivideogpt,li2025latent,feng2025vidarc, chen2024diffusion,huang2025self,bjorck2025gr00t}.
A common approach treats video generation as a planning substrate: UniPi~\citep{du2023learning} generates future frames using video diffusion and recovers actions via inverse dynamics, while Uni-Sim~\citep{yang2023learning} and Genie~\citep{bruce2024genie} leverage world models to synthesize interaction data for downstream policy learning.
These methods demonstrate strong generalization across tasks and environments, but operate purely in pixel space and provide no mechanism to enforce geometric consistency in predicted futures.
Recent work addresses this limitation by incorporating explicit geometric structure.
TesserACT~\citep{zhen2025tesseract} jointly generates RGB, depth, and surface normals to enable spatiotemporal reconstruction and action prediction.
3DFlowAction~\citep{zhi20253dflowaction} represents actions as 3D flow fields, grounding planning in scene geometry.
WristWorld~\citep{qian2025wristworld} synthesizes wrist-view observations through reconstruction, while RoboTransfer~\citep{liu2025robotransfer} enforces multi-view geometric constraints via cross-view feature interactions.
Related approaches impose geometric consistency through pointmap alignment or multi-view supervision~\citep{liu2025geometry}, and multi-modal world models such as iMoWM~\citep{zhang2025imowm} and MinD~\citep{chi2025mind} jointly model video and action generation.
Despite their effectiveness, these approaches share a common limitation: they \emph{modify the output space} of the video model to explicitly predict geometric quantities (e.g., depth, normals, or flow), which requires large-scale geometric annotations and constrains the expressive capacity of pretrained video backbones.
In contrast, \ourmethod{} does not alter the output space.
Instead, it distills geometric structure into the model's \emph{internal representations} during training, enabling correspondence-aware generation while retaining a single-stream architecture with zero additional inference cost.

\subsection{Feed-Forward 3D and 4D Geometry Models}

\noindent Feed-forward geometry models infer dense 3D structure directly from visual observations, avoiding iterative optimization.
DUSt3R~\citep{wang2024dust3r} introduced this paradigm by predicting pixel-aligned pointmaps from image pairs, enabling unconstrained stereo reconstruction.
MASt3R~\citep{leroy2024grounding} improves geometric fidelity through local feature matching, while Fast3R~\citep{yang2025fast3r} scales to multi-view reconstruction via global fusion. Spann3R~\citep{wang2024spann3r} introduces a spatial memory mechanism for incremental reconstruction, and and $\pi^3$~\citep{wang2025pi3} further removes reference-view bias through permutation-equivariant visual geometry learning.
Extending these methods to dynamic scenes introduces additional challenges, as both camera and object motion must be disentangled. 
MonST3R~\citep{zhang2025monst3r} estimates temporally consistent pointmaps and recovers global geometry and camera poses via optimization, while DAS3R~\citep{xu2024das3r} accelerates this process.
CUT3R~\citep{wang2025continuous} enables feed-forward dynamic reconstruction through large-scale training, and Easi3R~\citep{chen2025easi3r} explores training-free generalization.
PAGE-4D~\citep{zhou2026page} further improves robustness by disentangling static and dynamic components via motion-aware masking.
These models are typically used as standalone perception modules for geometry estimation.
In contrast, \ourmethod{} repurposes them as \emph{correspondence teachers}: their learned representations encode depth, camera motion, and scene dynamics, providing a supervision signal that enforces geometrically consistent inter-frame correspondences within a video world model.
To our knowledge, this is the first work to use 4D geometry foundation models as training-time regularizers for video generation, rather than as direct predictors of geometric outputs.

%% file: section/method.tex
\section{\ourmethod{}}
\label{sec:method}

\begin{figure*}[t]
\begin{center}
\includegraphics[width=\linewidth]{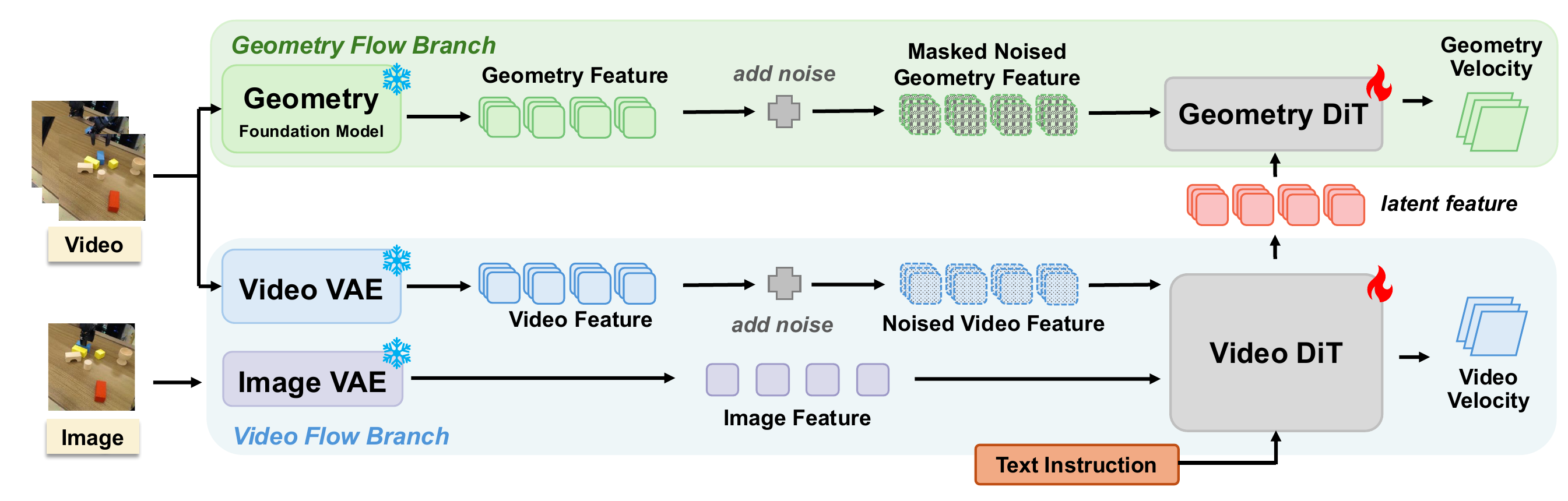}
\end{center}
\vspace{-0.2cm}
\caption{\textbf{\ourmethod{} training.} During training, a video DiT predicts the velocity of the noised video latent, while its intermediate features guide a geometry DiT to predict geometry velocity. This coupled training enforces geometry-consistent generation. During inference, only the video branch is used for efficient generation.}
\label{fig:pipeline}
\end{figure*}

\subsection{Problem Formulation}
\label{sec:problem}

\noindent Given an initial observation $\mathbf{I}_0$ and a language instruction $c$, we learn a world model
$M : (\mathbf{I}_0, c) \rightarrow \{\mathbf{I}_t\}_{t=1}^{N}$
that predicts future frames, and design a policy extraction module
$P : \{\mathbf{I}_t\}_{t=0}^{N} \rightarrow \{\mathbf{a}_t\}_{t=0}^{N-1}$
that extracts executable actions from the predicted rollout.
For $P$ to succeed, the generated rollout must preserve \emph{inter-frame correspondences}: pixels depicting the same 3D surface point must evolve consistently across time.
When this property is violated---even if the video appears photorealistic---action extraction becomes unreliable because the underlying 3D structure no longer reflects physically realizable motion~\cite{bharadhwaj2024track2act, wen2024atm}.
We enforce correspondence consistency during training via \textbf{Geometry-Enhanced Velocity Alignment} (Sec.~\ref{sec:gam}) shown in Fig.~\ref{fig:pipeline}, and convert correspondence-consistent rollouts into actions via an \textbf{Inverse Dynamic System} (Sec.~\ref{sec:ids}).

\subsection{Geometry-Enhanced Velocity Alignment}
\label{sec:gam}

\subsubsection{What Governs Inter-Frame Correspondence}
\label{sec:gam1}
We begin by formalizing what determines whether two pixels in adjacent frames correspond to the same physical point.
Let $\mathbf{X}_t \in \mathbb{R}^3$ be a scene point observed at pixel $\mathbf{p}_t$ in frame $t$. Under relative camera motion $(\mathbf{R}_{t \to t+1}, \mathbf{T}_{t \to t+1})$ and scene flow $\Delta \mathbf{X}_t$, its
projection $\mathbf{p}_{t+1}$ in frame $t{+}1$ is:
\vspace{1.2em}
\begin{figure}[H]
\centering
\resizebox{\columnwidth}{!}{%
\begin{minipage}{1.0\columnwidth}
\begin{equation}
\label{eq:correspondence}
\eqnmarkbox[Emerald]{pt1}{\mathbf{p}_{t+1}}
\sim
\eqnmarkbox[OliveGreen]{Kmat}{\mathbf{K}}
\Big[
\eqnmarkbox[NavyBlue]{R}{\mathbf{R}_{t\rightarrow t+1}}\,
\eqnmarkbox[WildStrawberry]{dpt}{\mathbf{D}(\mathbf{p}_t)}\,
\eqnmarkbox[OliveGreen]{Kinv}{\mathbf{K}^{-1}}\,
\eqnmarkbox[BurntOrange]{pt}{\mathbf{p}_t}
\;+\;
\eqnmarkbox[Plum]{T}{\mathbf{T}_{t\rightarrow t+1}}
\;+\;
\eqnmarkbox[Mahogany]{flow}{\Delta \mathbf{X}_t}
\Big].
\end{equation}
\vspace{1.2em}
\annotate[yshift=0.6em]{above,left}{pt1}{Pixel in next frame}
\annotate[yshift=0.6em]{above,right}{Kmat}{Intrinsic matrix}
\annotate[yshift=-1.8em]{below,right}{Kinv}{Inverse intrinsic matrix}
\annotate[yshift=-0.6em]{below,right}{pt}{Pixel in current frame}
\annotate[yshift=0.6em]{above,left}{T}{Camera translation}
\annotate[yshift=-0.6em]{below,left}{R}{Camera rotation}
\annotate[yshift=-1.8em]{below,left}{dpt}{Depth at $\mathbf{p}_t$}
\annotate[yshift=1.8em]{above,left}{flow}{Scene flow / object motion}
\end{minipage}%
}
\end{figure}

\noindent
This leads to two immediate insights.
First, pixel-level reconstruction losses cannot enforce correspondences: the mapping from scene geometry to pixel values is many-to-one, so different configurations of $(\mathbf{D}, \mathbf{R}, \mathbf{T}, \Delta \mathbf{X})$ can produce visually indistinguishable frames. 
A pixel loss can reach zero while the underlying correspondences are entirely wrong.
Second, a model whose internal representations correctly encode $(\mathbf{D}, \mathbf{R}, \mathbf{T}, \Delta \mathbf{X})$ \emph{necessarily} produces correct
correspondences, since Eq.~\ref{eq:correspondence} leaves no remaining degree of freedom.

\paragraph{Geometry foundation models as correspondence encoders.}
Models such as PAGE-4D~\citep{zhou2026page}, Depth Anything V3~\cite{lin2025depth},VGGT~\cite{wang2025vggt},and DUSt3R-family models~\cite{wang2024dust3r, zhang2025monst3r, wang2025continuous} employ cross-frame transformers supervised on two tasks: dense depth estimation and camera pose regression.
These are precisely the dominant factors in
Eq.~\ref{eq:correspondence}: given per-frame depth $\mathbf{D}$ and relative camera pose $(\mathbf{R}, \mathbf{T})$, the correspondence of any static scene point is fully determined. 
For dynamic points, the model must additionally capture how depth changes across frames---which implicitly encodes
object motion $\Delta \mathbf{X}$, since a point whose depth evolves inconsistently with the estimated camera motion must be moving independently. 
The learned feature representations of these models
therefore encode the complete correspondence structure of
Eq.~\ref{eq:correspondence}, despite being supervised only on depth and pose. 
Supervising the video backbone to predict these representations implicitly encourages correspondence consistency, without requiring an explicit correspondence loss.
GEM-4D shares REPA’s~\cite{yu2025repa} philosophy of leveraging foundation-model supervision to improve generative representations. Unlike REPA, which explicitly aligns the diffusion representation with foundation-model features, GEM-4D uses geometry foundation models as auxiliary supervision to encourage a shared video representation that jointly supports RGB denoising and 3D geometric reasoning.

\subsubsection{Flow Matching for Latent Video Generation}

We adopt flow matching~\citep{lipman2022flow} for latent video generation.
Let $\mathbf{z}_0$ be the VAE-encoded video latent~\cite{kingma2013auto} and $\mathbf{z}_1 \sim \mathcal{N}(\mathbf{0}, \mathbf{I})$ be noise.
A velocity field $\mathbf{v}_\theta^{\text{vid}}$ transports $\mathbf{z}_1$ to $\mathbf{z}_0$ via the ODE $d\mathbf{z}_t / dt = \mathbf{v}_\theta^{\text{vid}}(\mathbf{z}_t, t, c)$, trained by regressing against an analytically derived target velocity $\mathbf{v}^*$:
\begin{equation}
\mathcal{L}_{\mathrm{FM}}^{\text{vid}}
= \mathbb{E}_{\mathbf{z}_0, \mathbf{z}_1, t}\!\left[\,\|\mathbf{v}_\theta^{\text{vid}}(\mathbf{z}_t, t, c) - \mathbf{v}^*(\mathbf{z}_t, t)\|_2^2\,\right].
\label{eq:flow_loss}
\end{equation}
We parameterize the velocity through a Video DiT ~\cite{peebles2023scalable} with backbone $E_\theta^{\text{vid}}$ and output head $U_\theta^{\text{vid}}$:
\begin{equation}
\mathbf{m}_t = E_\theta^{\text{vid}}(\mathbf{z}_t, t, c), \qquad
\mathbf{v}_\theta^{\text{vid}} = U_\theta^{\text{vid}}(\mathbf{m}_t),
\label{eq:video_dit}
\end{equation}
where $\mathbf{m}_t$ denotes the intermediate features extracted at a mid-level layer.
Under standard flow matching, $\mathbf{m}_t$ is optimized only for appearance transport.
The next section describes how we shape $\mathbf{m}_t$ to encode correspondence structure.

\subsubsection{Correspondence Distillation via Geometry Flow}
Given a video sequence $\{\mathbf{I}_t\}_{t=0}^{T}$, a frozen geometry model $G$ extracts a dense geometric representation:
\begin{equation}
\mathbf{g}_0 = G\!\left(\{\mathbf{I}_t\}_{t=0}^{T}\right) \in \mathbb{R}^{T \times \frac{H}{P} \times \frac{W}{P} \times C}.
\label{eq:geo_repr}
\end{equation}
As established in Sec.~\ref{sec:gam1}, $\mathbf{g}_0$ encodes the factors $(\mathbf{D}, \mathbf{R}, \mathbf{T}, \Delta \mathbf{X})$ in Eq.~\ref{eq:correspondence}---it is a dense correspondence representation of the input video.
$G$ remains frozen throughout; it serves as a correspondence teacher whose knowledge we distill into the video backbone.
To perform this distillation, we introduce a parallel flow-matching process over the geometry representation space.
A Geometry DiT $\mathbf{v}_\psi^{\text{geo}}$, \emph{conditioned on the video backbone's intermediate features $\mathbf{m}_t$}, predicts the velocity field of the geometry latent:
\begin{equation}
\mathcal{L}_{\mathrm{FM}}^{\text{geo}}
= \mathbb{E}_{\mathbf{g}_0, \mathbf{g}_1, t}\!\left[\,\|\mathcal{M}(\mathbf{v}_\psi^{\text{geo}}(\mathbf{g}_t, t, \mathbf{m}_t)) - \mathbf{v}^*(\mathbf{g}_t, t)\|_2^2\,\right],
\label{eq:geo_loss}
\end{equation}
where $\mathbf{g}_t$ interpolates between $\mathbf{g}_0$ and noise $\mathbf{g}_1 \sim \mathcal{N}(\mathbf{0}, \mathbf{I})$; $\mathcal{M}$ denotes mask mechanism applied to noised geometry feature.
The critical design choice is that the Geometry DiT receives $\mathbf{m}_t$ as its \emph{only} scene-level conditioning signal.
It has no direct access to pixels, camera parameters, or depth maps---all scene information must arrive through $\mathbf{m}_t$.
Minimizing $\mathcal{L}_{\mathrm{FM}}^{\text{geo}}$ therefore requires $\mathbf{m}_t$ to contain sufficient information about the geometric factors $(\mathbf{D}, \mathbf{R}, \mathbf{T}, \Delta \mathbf{X})$ to predict how the geometry representation evolves over time.
Since these factors determine correspondences (Eq.~\ref{eq:correspondence}), the geometry loss could serve as a correspondence loss imposed on the video backbone's internal representations.

\subsubsection{Joint Objective and Geometric Grounding}

The training objective is:
\begin{equation}
\mathcal{L} = \mathcal{L}_{\mathrm{FM}}^{\text{vid}}
+ \alpha\,\mathcal{L}_{\mathrm{FM}}^{\text{geo}},
\label{eq:joint}
\end{equation}
where $\mathbf{\alpha}$ balances the two terms. Because both losses share the intermediate representation $\mathbf{m}_t$, the gradient with respect to the video backbone parameters $\theta$ decomposes as:
\vspace{1.6em}
\begin{equation}
\label{eq:gradient}
\eqnmarkbox[Emerald]{grad}{\nabla_\theta \mathcal{L}}
=
\eqnmarkbox[NavyBlue]{vidgrad}{
\nabla_\theta \mathcal{L}_{\mathrm{FM}}^{\text{vid}}}
\;+\;
\eqnmarkbox[Plum]{alpha}{\alpha}\cdot
\eqnmarkbox[WildStrawberry]{chain}{
\frac{\partial \mathcal{L}_{\mathrm{FM}}^{\text{geo}}}{\partial \mathbf{m}_t}
\cdot
\frac{\partial \mathbf{m}_t}{\partial \mathbf{\theta}}}
\end{equation}
\annotate[yshift=0.6em]{above,left}{grad}{Total gradient}
\annotate[yshift=-0.8em]{below,left}{vidgrad}{Appearance supervision}
\annotate[yshift=0.6em]{above,left}{chain}{Geometry-induced gradient}
\annotate[yshift=-0.8em]{below,right}{alpha}{Balancing weight}
\vspace{0.9em}

Because both losses share the intermediate representation $\mathbf{m}_t$, the gradient decomposes into an appearance term and a geometry-induced term. 
The latter propagates correspondence structure into the video backbone through $\mathbf{m}_t$.
The first term drives $\mathbf{m}_t$ to encode how pixels move over
time---appearance transport. The second, which is non-zero by
construction since $\mathcal{L}_{\mathrm{FM}}^{\text{geo}}$ depends on
$\mathbf{m}_t$ (Eq.~\ref{eq:video_dit}), drives $\mathbf{m}_t$ to additionally encode
\emph{why} they move that way: the depth, camera pose, and scene flow
that determine correspondences (Eq.~\ref{eq:correspondence}). At
convergence, $\mathbf{m}_t$ must satisfy both objectives, eliminating
representations that explain appearance but violate geometric
consistency.

\subsection{Adaptive Inverse Dynamic System} 
\label{sec:ids}

\begin{figure*}[t]
\begin{center}
\includegraphics[width=\linewidth]{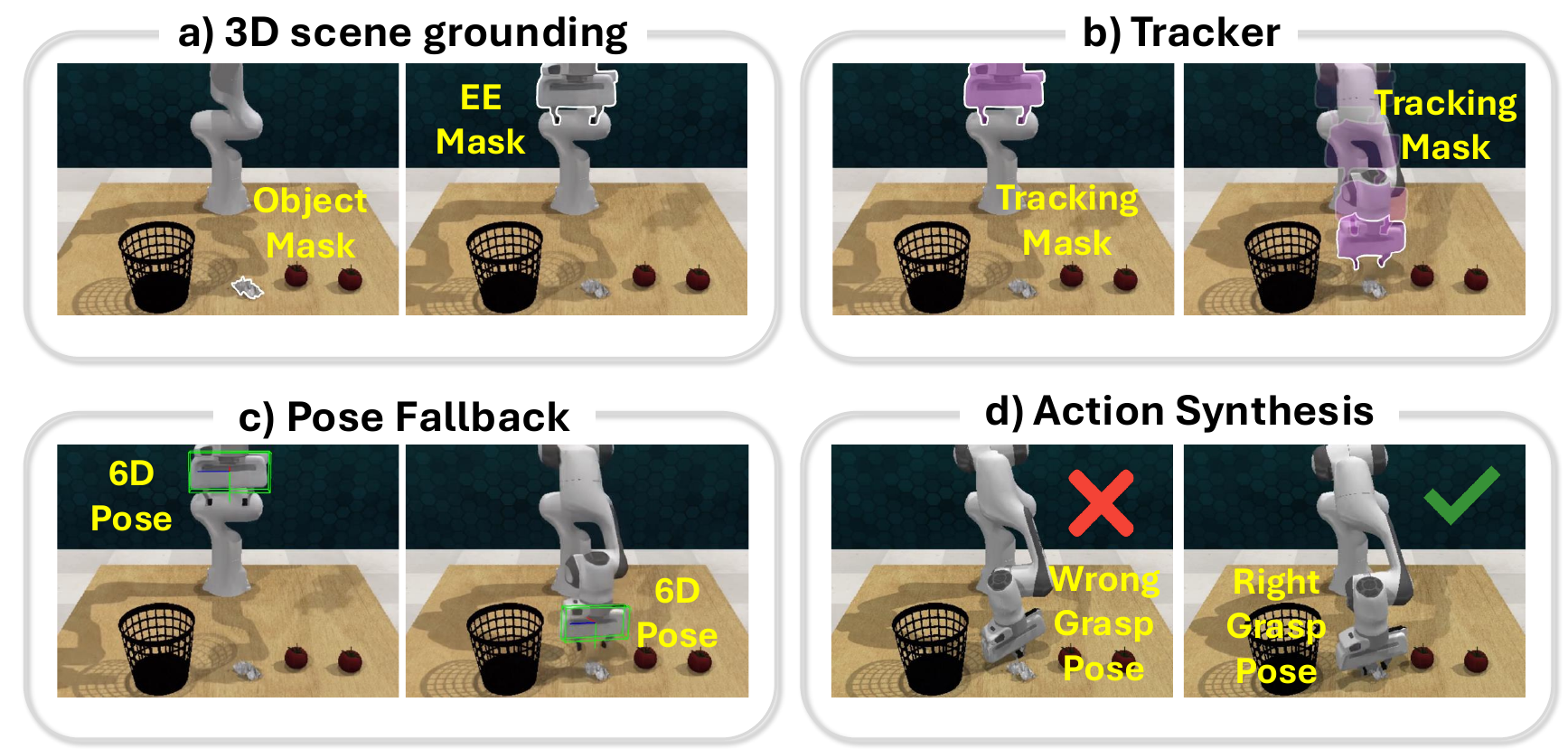}
\end{center}
\vspace{-0.2cm}
\caption{\textbf{Adaptive Inverse Dynamic System.} Given a generated video as input, this system extracts a robot policy through the four steps illustrated in the figure.} 
\label{fig:AIDS_pipline}
\end{figure*}

\noindent Given generated frames $\{\mathbf{I}_t\}_{t=0}^{N}$, we make use of \textbf{Adaptive Inverse Dynamic System (AIDS)} to convert it into executable 6-DoF action trajectories 
$\{\mathbf{a}_t\}_{t=0}^{N-1}$, as illustrated in \cref{fig:AIDS_pipline}.
AIDS introduces two mechanisms that together make action extraction robust to these artifacts in generated results without any task-specific training: (i) a \emph{dual-criterion confidence-gated tracker} that separates gradual drift from catastrophic collapse and invokes heavyweight VLM re-grounding only when warranted, and (ii) a \emph{geometry--kinematics pose fallback} that decouples translation (well-observed from depth) from rotation (temporally smoothed in $SE(3)$) whenever learned pose estimation becomes unreliable.

\paragraph{3D scene grounding.}
We first localize the target object and end-effector (EE) in 3D. 
Given the instruction, the depth map, and camera intrinsics (estimated by the geometry foundation model), Qwen3.5-VL~\cite{wu2025qwen} and SAM 2~\cite{ravi2024sam, kirillov2023segment} generate segmentation masks for the target object and the end effector (EE), which are then used to extract their corresponding point clouds.
We then align the EE CAD to the EE point cloud with FoundationPose~\cite{wen2024foundationpose} to recover the initial EE translation and rotation $\left(\mathbf{R}_{\text{ee}}^{0}, \mathbf{T}_{\text{ee}}^{0}\right) \in SE(3)$.

\paragraph{Dual-Criterion Confidence-Gated Tracker.}
With initial EE pose, we propagate dense keypoints sampled from the mask of end effector $\mathcal{M}_{\text{ee}}$ through the rollout using CoTracker3~\cite{karaev2025cotracker3, karaev2024cotracker}. Let $\mathcal{V}_{t_0}\subseteq\mathcal{M}_{\text{ee}}$ be the anchor keypoint set at initial frame, and $\mathcal{V}_t\subseteq\mathcal{V}_{t_0}$ the subset still reliably tracked at frame $t$. We monitor following two metrics:
\begin{equation}
s_t = \frac{|\mathcal{V}_t|}{|\mathcal{V}_{t_0}|} \in [0,1], \qquad \Delta s_t = s_t - s_{t-1},
\end{equation}
where $s_t$ denotes the anchor retention ratio ($s_t=1$ means all anchors remain reliably tracked, $s_t{\to}0$ means tracker failure) and $\Delta s_t$ denotes the frame-to-frame change (negative value signals a loss of correspondence).
These two signals separate two qualitatively different failure modes. A \emph{gradual drift} manifests as $s_t$ decaying smoothly below the retention threshold $\tau\in(0,1)$; an \emph{abrupt collapse}, typically caused by frame-level generative artifacts, manifests as a sharp drop $\Delta s_t<-\delta$, where $\delta>0$ is the drop threshold.
We handle these two failure modes with different interventions:
\begin{equation}
\hat{\mathcal{M}}_{\text{ee}}^{\,t} =
\begin{cases}
\text{re-anchor tracker at } t, & \text{if } s_t < \tau,\\[2pt]
\mathrm{Qwen3.5\text{-}VL}(I_t,\,c), & \text{if } \Delta s_t < -\delta,\\[2pt]
\mathcal{M}_{\text{ee}}^{\,t}, & \text{otherwise.}
\end{cases}
\end{equation}

The first branch corresponds to a \emph{gradual drift} (resampling fresh keypoints from the most recent reliable mask), and the second to an \emph{abrupt collapse} (re-grounding the EE mask itself via vision-language semantics).

\begin{figure*}[t]
\begin{center}
\includegraphics[width=\linewidth]{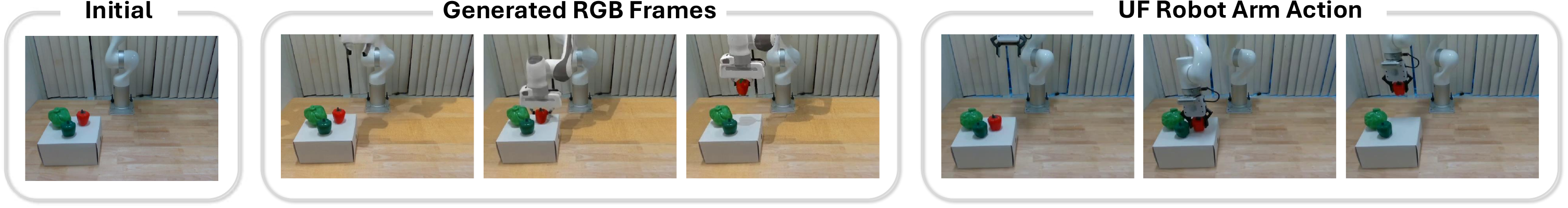}
\end{center}
\vspace{-0.2cm}
\caption{\textbf{Generated Frames to Arm Action.} Starting from an initial observation, \ourmethod{} predicts future frames, which are then converted into executable UF arm actions. The generated frames are displayed with higher brightness to distinguish them from the real frames.} 
\vspace{-0.6cm}
\label{fig:real_world_result}
\end{figure*}
\paragraph{Geometry–kinematics pose fallback.}
Given the mask of end effector $\mathcal{M}_{\text{ee}}^{\,t}$, the frame $\mathbf{I}_t$ , the depth $\mathbf{D}_t$, and the EE CAD model, FoundationPose~\cite{wen2024foundationpose} predicts the pose of EE $\left(\mathbf{R}_{\text{ee}}^{t}, \mathbf{T}_{\text{ee}}^{t}\right)$ together with a confidence $\kappa_t\in[0,1]$ for each frame.

\begin{equation}
(\mathbf{R}_{\text{ee}}^{t},\mathbf{T}_{\text{ee}}^{t}, \kappa_t) = \mathrm{FoundationPose}\bigl(\mathbf{I}_t,\mathbf{D}_t,\mathcal{M}_{\mathrm{ee}}^{\,t},\mathrm{CAD}\bigr),
\end{equation}

When $\kappa_t < \kappa^\ast$, where $\kappa^\ast\in[0,1]$ is a FoundationPose acceptance confidence threshold, we apply a temporal consistency check and reject the FoundationPose estimate if either the translation jump or the rotation jump exceeds its corresponding threshold: 
\begin{equation}
\|\mathbf{T}_{\mathrm{ee}}^t - \mathbf{T}_{\mathrm{ee}}^{t-1}\|_2 > \epsilon_t, \quad d_{\mathrm{geo}}(\mathbf{R}_{\mathrm{ee}}^t, \mathbf{R}_{\mathrm{ee}}^{t-1}) > \epsilon_R.
\end{equation}

Where $d_{\mathrm{geo}}(\cdot,\cdot)$ denotes the geodesic distance on $SO(3)$. For rejected frames, we recover the EE translation by back-projecting valid depth pixels within the EE mask and taking their 3D centroid, while the missing rotation is estimated by spherical linear interpolation~\cite{shoemake1985animating} between the nearest accepted poses in time. Finally, we'll have the recovered EE trajectory $\{(\mathbf{R}_{\mathrm{ee}}^{\,t},\mathbf{T}_{\mathrm{ee}}^{\,t})\}_{t=0}^{N}$.

\paragraph{Grasp insertion and action synthesis.}
From the recovered EE trajectory, we first select the pose closest to the target object as a reference pose, denoted by $(\mathbf{R}_{\mathrm{ref}},\mathbf{T}_{\mathrm{ref}})$. GraspGen~\cite{murali2025graspgen} then predicts a set of grasp candidates $\{(\mathbf{R}_{\mathrm{grasp}}^i,\mathbf{T}_{\mathrm{grasp}}^i)\}_{i=0}^{M}$ on the target object point cloud. We rank these candidates by their weighted pose deviation from reference pose, combining translation and rotation errors, and select the best-scoring candidate as the final grasp pose:
\begin{equation}
\mathbf{T}_{\mathrm{grasp}}^\ast
=
\arg\min_{\mathbf{T}_{\mathrm{grasp}}^{(i)}}
\Bigl(
\lambda_t \, \|\mathbf{t}_{\mathrm{grasp}}^{(i)}-\mathbf{t}_{\mathrm{ref}}\|_2
+
\lambda_R \, d_{\mathrm{geo}}\!\bigl(\mathbf{R}_{\mathrm{grasp}}^{(i)},\mathbf{R}_{\mathrm{ref}}\bigr)
\Bigr),
\end{equation}
where $\lambda_t$ and $\lambda_R$ balance translation and rotation consistency, respectively. We then insert $\mathbf{T}_{\mathrm{grasp}}^\ast$ into the recovered EE trajectory as an intermediate target and smooth the resulting motion via interpolation.
Inverse kinematics then converts the smoothed trajectory into an executable action sequence $\{\mathbf{a}_t\}_{t=1}^{N}$ for controlling the robot arm, as illustrated in \cref{fig:real_world_result}.

%% file: section/exp.tex
\section{Experiments}

\noindent \textbf{Overview.} 
Our experiments are structured around two research questions.
First, does geometry distillation improve 4D scene prediction in terms of appearance, depth, and correspondence consistency?
Second, can the resulting geometry-consistent rollouts improve downstream robot manipulation when converted into executable actions?
We therefore organize the evaluation into two parts: Sec.~\ref{sec:4dscene} studies 4D scene reconstruction quality. Sec.~\ref{sec:embodied} evaluates whether the generated rollouts support more reliable robot action planning and execution. Sec.~\ref{sec:ablation} examines the effectiveness of the proposed geometry flow branch.

\begin{figure*}[t]
\begin{center}
\includegraphics[width=\linewidth]{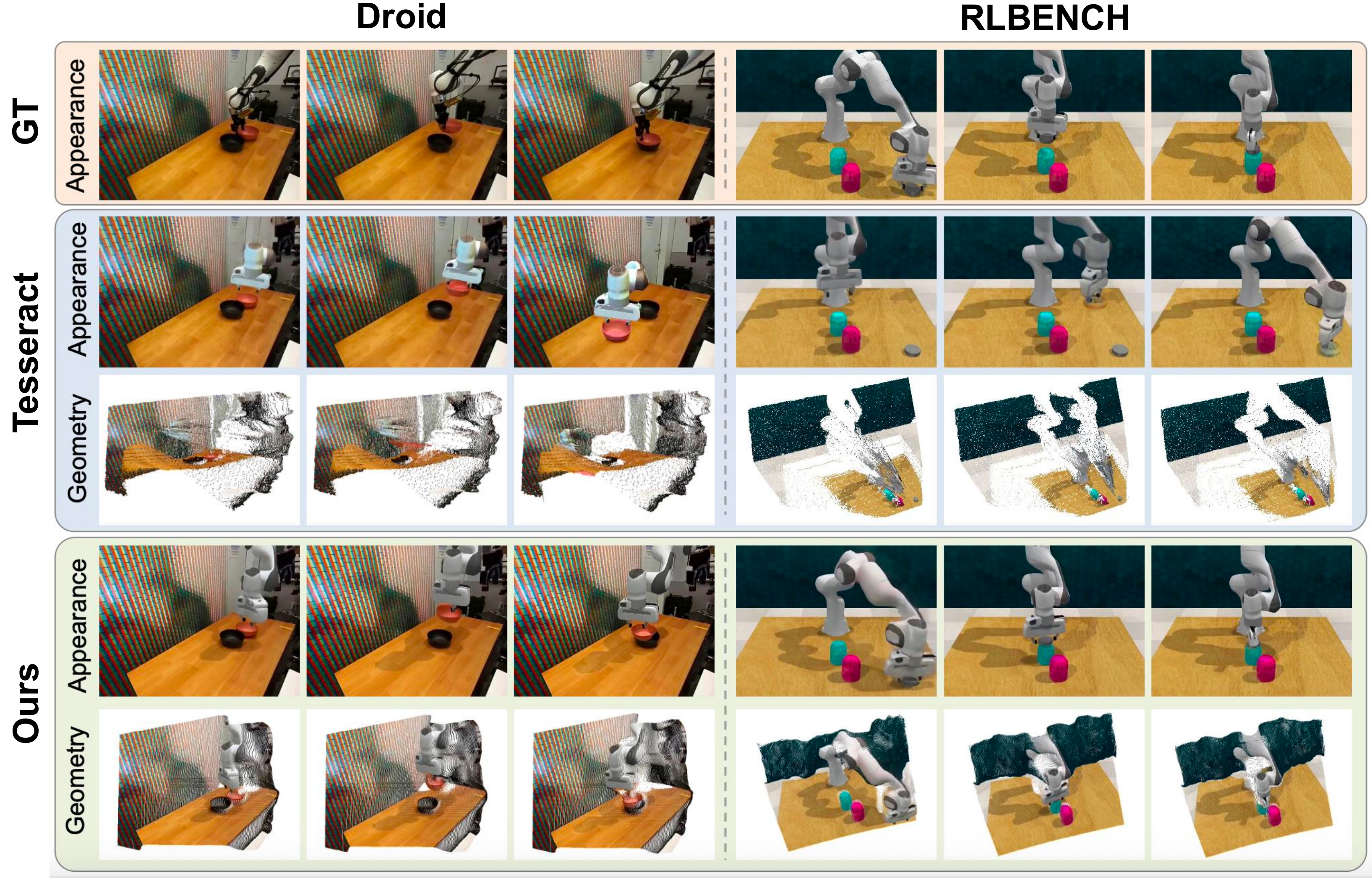}
\end{center}
\vspace{-0.2cm}
\caption{\textbf{Qualitative 4D scene generation results on Droid (real) and RLBench (synthetic).} 
Top: ground truth. Middle: TesserAct generates plausible RGB but produces inconsistent depth — note the warping on the manipulator and the noisy depth strip on Droid. Bottom: GEM-4D preserves geometric structure across frames, yielding cleaner depth and tighter object boundaries.} 
\vspace{-0.6cm}
\label{fig:result1}
\end{figure*}

\paragraph{Datasets.} 
We train our model on tasks simulated in ManiSkill3~\cite{tao2024maniskill3} and RLBench~\cite{james2020rlbench}, as well as on the Bridge~\cite{walke2023bridgedata} and RT-1~\cite{brohan2022rt} datasets. 
To evaluate both video generation quality and robotic modeling capability, we conduct experiments in real-world and synthetic domains. 
In the real-world setting, we use 400 unseen samples from the Droid dataset~\cite{khazatsky2024droid}, where depth is estimated using Depth Anything V3~\cite{lin2025depth} and the point tracking is estimated using CoTracker3~\cite{karaev2025cotracker3}.
In the synthetic setting, we evaluate on 780 unseen samples from RLBench~\cite{james2020rlbench}, where ground-truth depth is directly available and the point tracking is estimated using CoTracker3~\cite{karaev2025cotracker3}.

\paragraph{Baselines.} 
Our method is compared against the following: 
(1) \textbf{CogVideoX}~\cite{yang2024cogvideox}, a large-scale image-to-video generation model based on a diffusion transformer architecture, pretrained on approximately 35M video clips. We fine-tune it on our training set. 
(2) \textbf{WAN 2.2-14B}~\cite{wan2025wan}, a large-scale image-to-video generation model built upon a spatio-temporal variational autoencoder. We fine-tune it on our training set. 
(3) \textbf{TesserAct}~\cite{zhen2025tesseract}, a CogVideoX-based 4D embodied world model that further fine-tunes CogVideoX on robotics datasets while jointly modeling depth and surface normals.
(4)
\textbf{Geometry Forcing}~\cite{wu2025geometry}, a representation-alignment-based method that incorporates geometric prior knowledge into the generative model through feature alignment.
We do not compare directly against RoboTransfer~\cite{liu2025robotransfer} (multi-view RGB with depth/normal conditioning), 3DFlowAction~\cite{zhi20253dflowaction} (predicts 3D flow rather than RGB), or Liu et al.~\cite{liu2025geometry} (stereo RGB-D with cross-view pointmap supervision), as their input modalities and output spaces differ from our single-view RGB-only setting.

\subsection{4D Scene Prediction}
\label{sec:4dscene}

\noindent\textbf{Metrics.}
We evaluate video generation quality using FVD~\cite{unterthiner2019fvd}, SSIM~\cite{wang2004image}, and PSNR~\cite{zhou2023dynpoint}; depth estimation accuracy using AbsRel, $\delta_1$, and $\delta_2$
\cite{eigen2014depth}.
To assess 3D reconstruction quality, we compute the L1 Chamfer Distance
\cite{fan2017point, zhou2026rise}
between the predicted and ground-truth point clouds. 
\input{table/video_syn}
The correspondence quality is measured by using $\delta^{vis}_{avg}$ (fraction of visible points tracked within 1, 2, 4, 8 and 16 pixels, averaged over thresholds)
\cite{doersch2022tapvid}.
Each sample is generated 20 times, and we report the averaged results.

\paragraph{Results and analysis.}
Table~\ref{tab:4d_scene_results} reports quantitative comparisons on both real and synthetic datasets. 
Qualitative results are shown in Fig~\ref{fig:result1}.
Our method consistently achieves the best overall performance across most RGB appearance metrics and depth reconstruction metrics. 
In particular, \ourmethod{} achieves the lowest FVD and the highest SSIM/PSNR on the real-world dataset, indicating that it generates videos with higher visual quality and temporal consistency than the baseline models.
More importantly, when evaluating geometric quality using DA3~\cite{lin2025depth} for depth estimation, our method significantly outperforms existing baselines. 
On the real dataset, \ourmethod{} achieves the lowest AbsRel and the highest $\delta_1$ and $\delta_2$, demonstrating that the generated frames preserve more accurate scene geometry across time. 
A similar trend is observed on the synthetic dataset, where our method produces the most accurate depth reconstruction.
These results suggest that the proposed geometry-aware training effectively improves both appearance generation and geometric consistency. 
By incorporating geometry dynamics during training, the model learns to generate videos that not only look realistic but also maintain coherent 3D structure, which further leads to improved downstream reconstruction quality~\cite{huang2026pointworld} when processed by DA3.

\input{table/action}

\subsection{Embodied Action Planning}
\label{sec:embodied}

\subsubsection{Quantitative Experiment}
\textbf{Dataset.}
We evaluate our model on 7 challenging manipulation tasks from RLBench~\cite{james2020rlbench} and unseen realistic dataset from Droid~\cite{khazatsky2024droid}.
These tasks are selected for their requirements of precise grasping and complex spatial reasoning.
\textbf{Metric.}
To ensure a fair evaluation, we adopt two complementary evaluation protocols. 
For realistic dataset, we conduct a human study to assess the quality of generated videos, focusing on task success, object and robot-arm deformation, and instruction following. (Averaged over 15 participants; more generated results are provided at the link in the abstract.)
For simulation dataset, we convert generated videos into executable action trajectories, and measure task success rate by replaying the resulting trajectories in the simulator.

\textit{Realistic Dataset.}
Human evaluation on real-world Droid tasks demonstrates that the benefits of \ourmethod\ extend well beyond simulation (Tab.~\ref{tab:action}). \ourmethod\ consistently achieves substantially higher task success across all benchmarks, improving performance from $58\%\!\rightarrow\!75\%$ on AUTOLab, $65\%\!\rightarrow\!83\%$ on CLVR, and $59\%\!\rightarrow\!87\%$ on RAIL. The consistent gains across diverse real-world environments indicate that geometry-aware supervision yields more robust and transferable world representations for long-horizon manipulation.
\textit{Simulation Dataset.}
The right part of Tab.~\ref{tab:action} further evaluates policy extraction by re-executing predicted trajectories in simulation. 
Compared with TesserAct and CogVideoX, GEM-4D achieves higher planning success across most tasks. 
As the inverse dynamics module of TesserAct is not open-source, we use our AIDS to process its generated videos.
Because most videos generated by CogVideoX cannot be reliably processed by our AIDS, we do not report task success rates for this baseline.
\begin{figure*}[t]
\begin{center}
\includegraphics[width=\linewidth]{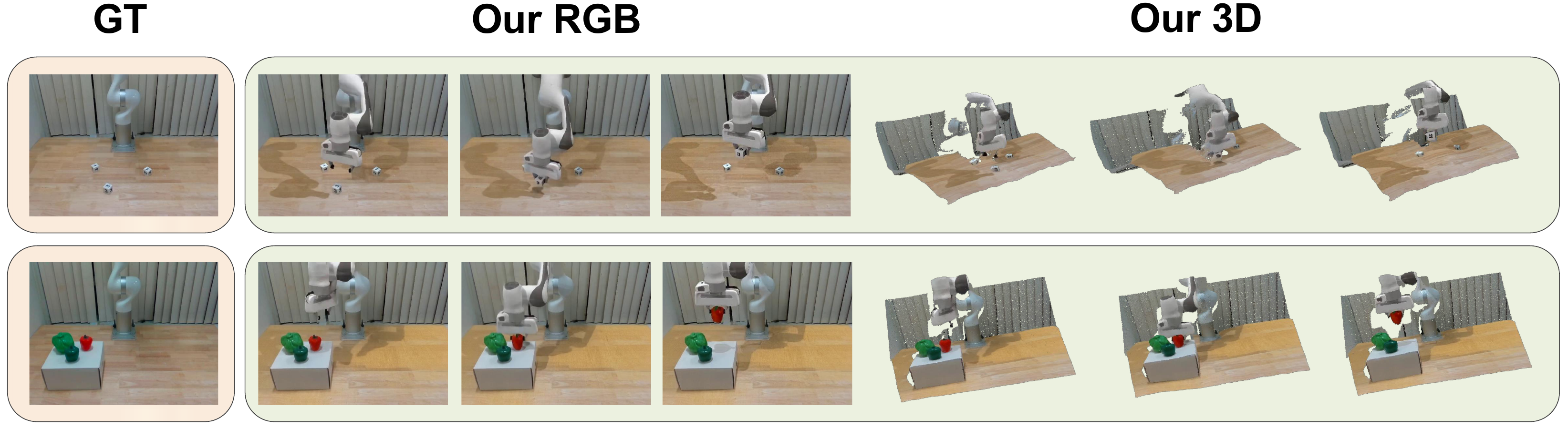}
\end{center}
\vspace{-0.2cm}
\caption{\textbf{Real-robot rollouts.} From left: ground-truth images, GEM-4D-generated RGB, and the back-projected 3D point cloud. The model produces realistic and geometrically coherent rollouts supporting transfer to UF Arm manipulation.} 
\vspace{-0.3cm}
\label{fig:result2}
\end{figure*}
\input{table/ablation}

While TesserAct frequently fails to produce executable trajectories, GEM-4D reaches $63\%$–$82\%$ success on RLBench tasks and consistently improves performance on long-horizon manipulation benchmarks. 
This shows that geometry-consistent world modeling significantly improves downstream action planning.

\subsubsection{Qualitative Experiment}
To evaluate our model in realistic settings, we conducted several tasks using a UF robot arm in real-world scenarios. The results are shown in Fig.~\ref{fig:result2}. We observe that, even under unseen backgrounds, our method generates realistic visual results and reasonable geometry, demonstrating strong potential for transfer to real robot manipulation.

\subsection{Ablation Studies}
\label{sec:ablation}
\noindent
We conduct a series of ablation studies to examine the contributions of the proposed geometry flow branch and different geometry priors. The results are summarized in Table~\ref{tab:ablation}.
First, we consider a baseline obtained by directly fine-tuning Generation backbone without any geometry guidance, in order to test whether fine-tuning alone is sufficient to improve 4D generation. 
Second, we replace the proposed geometry representation with explicit depth supervision, training the model to predict depth velocity rather than geometry features (\textbf{\ourmethod\ (Dep)}). This setting evaluates whether direct depth constraints can provide comparable geometric guidance. 
Third, we compare different geometry foundation models, including VGGT, to study how the choice of geometry prior influences learning (\textbf{\ourmethod\ (VGGT)}). 
The results reveal several consistent trends. Directly incorporating VGGT features slightly degrades performance, which may stem from the fact that VGGT is primarily trained for static or quasi-static scenes and is therefore less well matched to the dynamic scene evolution required in robotic manipulation. 
Using depth as a geometric constraint, by contrast, yields competitive performance. 
Notably, unlike Tesseract, which conditions on first-frame depth and predicts future depth, our depth ablation receives only noise as input and learns depth through the flow-matching objective alone. 

%% file: table/video_syn.tex
\begin{table*}[t]
\centering
\caption{
\textbf{Quantitative comparison on 4D scene generation.}
We evaluate RGB reconstruction, depth reconstruction and points correspondence quality across both real and synthetic domains.
The best results are highlighted in \textbf{bold}, and the second-best results are \underline{underlined}. \textbf{D} denotes dataset; \textbf{R} denote realistic dataset and \textbf{S} denote simulation dataset. Std. and dev. across 20 generations is around $1\%-2\%$.}
\vspace{-0.3em}

\resizebox{\linewidth}{!}{
\begin{tabular}{llcccccccc}
\toprule
\textbf{D} & \textbf{Method} &
\multicolumn{3}{c}{\textbf{RGB}} &
\multicolumn{3}{c}{\textbf{Depth}} &
\multicolumn{1}{c}{\textbf{Points}} &
\multicolumn{1}{c}{\textbf{Tracking}}  \\
\cmidrule(r){3-5} \cmidrule(r){6-8} \cmidrule(r){9-9} \cmidrule(r){10-10} 
& &
\cellcolor{col2} FVD $\downarrow$ &
\cellcolor{col1} SSIM $\uparrow$ &
\cellcolor{col1} PSNR $\uparrow$ &
\cellcolor{col2} AbsRel $\downarrow$ &
\cellcolor{col1} $\delta_1$ $\uparrow$ &
\cellcolor{col1} $\delta_2$ $\uparrow$ &
\cellcolor{col2} Chamfer $\downarrow$ &
\cellcolor{col1} $\delta^{vis}_{avg}$ $\uparrow$ \\
\midrule

\multirow{4}{*}{\textbf{R}}
& CogVideoX~\cite{yang2024cogvideox} & 35.56 & 75.91 & 20.18 & 22.33 & 68.32 & 83.17 & 0.2670 & 66.22 \\
& Wan 2.2-14B~\cite{wan2025wan} & 33.43 & \underline{76.24} & \underline{20.70} & \underline{21.39} & \underline{71.18} & \underline{84.35} & \underline{0.2349}  & \underline{68.18}\\
& TesserAct~\cite{zhen2025tesseract} & 33.28 & 75.66 & 20.08 & 22.07 & 66.80 & 82.60 & 0.2630 & 67.14 \\
& Geometry-Forcing~\cite{wu2025geometry}   & \underline{33.17} & 76.12 & 20.53 & 21.96 & 69.74 & 83.83 & 0.2443 & 67.97\\
& \textbf{\ourmethod} & \textbf{31.82} & \textbf{82.05} & \textbf{21.11} & \textbf{20.13} & \textbf{78.19} & \textbf{88.21} & \textbf{0.2001} & \textbf{71.23}\\
\midrule

\multirow{4}{*}{\textbf{S}}
& CogVideoX~\cite{yang2024cogvideox} & 40.21 & 75.51 & \underline{20.03} & 15.41 & \underline{70.99} & 92.90 & 0.2913  & 58.32\\
& Wan 2.2-14B~\cite{wan2025wan} & 49.20 & 73.01 & 19.87 & 17.81 & 67.07 & 90.16 & 0.1762 & \underline{61.99} \\
& TesserAct~\cite{zhen2025tesseract} & 41.97 & 76.72 & 19.71 & 16.02 & 69.26 & \underline{93.03} & 0.1813  & 61.15\\
& Geometry-Forcing~\cite{wu2025geometry}   & \underline{34.06} & \underline{77.92} & 19.48 & \underline{15.34} & 68.96 & 92.80 & \underline{0.1488}& 60.84 \\
& \textbf{\ourmethod} & \textbf{27.94} & \textbf{80.27} & \textbf{23.36} & \textbf{14.11} & \textbf{74.13} & \textbf{95.01} & \textbf{0.0702} & \textbf{68.18} \\
\bottomrule
\end{tabular}
}
\vspace{-1.0em}
\label{tab:4d_scene_results}
\end{table*}

%% file: table/action.tex
\begin{table}[t]
\centering
\caption{\textbf{Quantitative comparison on task success rates.} Comparison of task success rates across Droid and RLBench tasks. For the real-world Droid dataset, success rates are measured through a human study. For the simulated RLBench dataset, success rates are measured by executing the generated trajectories in simulation.}
\label{tab:action}
\resizebox{\linewidth}{!}{
\begin{tabular}{lcccccccccc}
\toprule
& \multicolumn{3}{c}{Droid (Real)} & \multicolumn{7}{c}{RLBench (Simulator)} \\
\cmidrule(lr){2-4} \cmidrule(lr){5-11}
& \makecell[c]{AUTOLab}
& \makecell[c]{CLVR}
& \makecell[c]{RAIL}
& \makecell[c]{Lift\\Numbered\\Block}
& \makecell[c]{Put\\Rubbish\\InBin}
& \makecell[c]{Reach\\Target}
& \makecell[c]{Lamp\\On}
& \makecell[c]{Pick\\Up\\Cup}
& \makecell[c]{Slide\\Block\\ToTarget}
& \makecell[c]{Solve\\Puzzle} \\
\midrule
CogVideoX~\cite{yang2024cogvideox} & 49 & 64 & 39 & - & - & - & - & - & - & - \\
Tesseract~\cite{zhen2025tesseract} & \underline{58} & \underline{65} & \underline{59} & \underline{21} & \underline{0} & \underline{2} & \underline{36} & \underline{49} & \underline{18} & \underline{33} \\
Ours      & \textbf{75} & \textbf{83} & \textbf{87} & \textbf{78} & \textbf{75} & \textbf{82} & \textbf{67} & \textbf{81} & \textbf{80} & \textbf{63} \\
\bottomrule
\end{tabular}
}
\end{table}

%% file: table/ablation.tex
\begin{table*}[t]
\centering
\caption{
\textbf{Ablation studies on 4D scene generation.} We evaluate RGB
reconstruction, depth reconstruction and points correspondence quality on
real domain.}
\vspace{0.3em}

\resizebox{\linewidth}{!}{
\begin{tabular}{llccccccc}
\toprule
\textbf{Domain} & \textbf{Method} &
\multicolumn{3}{c}{\textbf{RGB (Appearance)}} &
\multicolumn{3}{c}{\textbf{Depth (Geometry)}} &
\multicolumn{1}{c}{\textbf{P (Geometry)}} \\
\cmidrule(r){3-5} \cmidrule(r){6-8} \cmidrule(r){9-9}
& &
\cellcolor{col2} FVD $\downarrow$ &
\cellcolor{col1} SSIM $\uparrow$ &
\cellcolor{col1} PSNR $\uparrow$ &
\cellcolor{col2} AbsRel $\downarrow$ &
\cellcolor{col1} $\delta_1$ $\uparrow$ &
\cellcolor{col1} $\delta_2$ $\uparrow$ &
\cellcolor{col2} Chamfer $\downarrow$ \\
\midrule

\multirow{5}{*}{Real}
& CogVideoX~\cite{yang2024cogvideox} & 35.56 & 75.91 & 20.18 & 22.33 & 68.32 & 83.17 & 0.2670 \\
& Wan 2.2-14B~\cite{wan2025wan}    & 33.43 & 76.24 & 20.70 & 21.39 & 71.18 & 84.35 & 0.2349 \\
& \ourmethod (Dep)  & \underline{32.91} & \underline{78.58} & \underline{20.75} & \underline{20.89} & \underline{74.60} & \underline{86.67} &  \underline{0.2229} \\
& \ourmethod (VGGT)~\cite{wang2025vggt}   & 33.68 & 75.89 & 20.64 & 21.73 &  71.03 & 83.80 & 0.2370\\
& \ourmethod       & \textbf{31.82} & \textbf{82.05} & \textbf{21.11} & \textbf{20.13} & \textbf{78.19} & \textbf{88.21} & \textbf{0.2001}  \\

\bottomrule
\end{tabular}
}

\vspace{-0.5em}
\label{tab:ablation}
\end{table*}

%% file: section/conclusion.tex
\section{Conclusion}
\noindent
We presented GEM-4D, a geometry-enhanced video world model for robot manipulation. Unlike prior video world models that often generate visually plausible but geometrically inconsistent futures, GEM-4D incorporates supervision from geometry foundation models to enforce correspondence-consistent scene evolution during training, while retaining a single-stream architecture with no additional inference cost.
Our approach improves both appearance generation and geometric fidelity, enabling predicted rollouts to support downstream action extraction. To close the loop from generation to control, we further introduced an adaptive inverse dynamics system that converts generated videos into executable robot trajectories. Experiments across real-world and simulated benchmarks demonstrate consistent gains in video prediction, geometric consistency, and manipulation success. Overall, our results suggest that distilling geometric structure into video world models is a simple and effective step toward more reliable world models for embodied AI.

\newpage

%% file: egbib.bib
@String(CVPR  = {IEEE Conf. Comput. Vis. Pattern Recog.})

@String(ECCV  = {Eur. Conf. Comput. Vis.})

@String(NeurIPS = {Adv. Neural Inform. Process. Syst.})

@String(ICLR  = {Int. Conf. Learn. Represent.})

@article{du2023learning,
  title={Learning universal policies via text-guided video generation},
  author={Du, Yilun and Yang, Sherry and Dai, Bo and Dai, Hanjun and Nachum, Ofir and Tenenbaum, Josh and Schuurmans, Dale and Abbeel, Pieter},
  journal={Advances in neural information processing systems},
  volume={36},
  pages={9156--9172},
  year={2023}
}

@article{yang2023learning,
  title={Learning interactive real-world simulators},
  author={Yang, Mengjiao and Du, Yilun and Ghasemipour, Kamyar and Tompson, Jonathan and Schuurmans, Dale and Abbeel, Pieter},
  journal={arXiv preprint arXiv:2310.06114},
  year={2023}
}

@article{du2023video,
  title={Video language planning},
  author={Du, Yilun and Yang, Mengjiao and Florence, Pete and Xia, Fei and Wahid, Ayzaan and Ichter, Brian and Sermanet, Pierre and Yu, Tianhe and Abbeel, Pieter and Tenenbaum, Joshua B and others},
  journal={arXiv preprint arXiv:2310.10625},
  year={2023}
}

@article{zhou2024robodreamer,
  title={RoboDreamer: Learning Compositional World Models for Robot Imagination},
  author={Zhou, Siyuan and Du, Yilun and Chen, Jiaben and Li, Yandong and Yeung, Dit-Yan and Gan, Chuang},
  journal={arXiv preprint arXiv:2404.12377},
  year={2024}
}

@article{bharadhwaj2024gen2act,
  title={Gen2act: Human video generation in novel scenarios enables generalizable robot manipulation},
  author={Bharadhwaj, Homanga and Dwibedi, Debidatta and Gupta, Abhinav and Tulsiani, Shubham and Doersch, Carl and Xiao, Ted and Shah, Dhruv and Xia, Fei and Sadigh, Dorsa and Kirmani, Sean},
  journal={arXiv preprint arXiv:2409.16283},
  year={2024}
}

@article{chi2024eva,
  title={EVA: An Embodied World Model for Future Video Anticipation},
  author={Chi, Xiaowei and Zhang, Hengyuan and Fan, Chun-Kai and Qi, Xingqun and Zhang, Rongyu and Chen, Anthony and Chan, Chi-min and Xue, Wei and Luo, Wenhan and Zhang, Shanghang and others},
  journal={arXiv preprint arXiv:2410.15461},
  year={2024}
}

@inproceedings{zhu2025aether,
  title={Aether: Geometric-aware unified world modeling},
  author={Zhu, Haoyi and Wang, Yifan and Zhou, Jianjun and Chang, Wenzheng and Zhou, Yang and Li, Zizun and Chen, Junyi and Shen, Chunhua and Pang, Jiangmiao and He, Tong},
  booktitle={Proceedings of the IEEE/CVF International Conference on Computer Vision},
  pages={8535--8546},
  year={2025}
}

@inproceedings{leroy2024grounding,
  title={Grounding image matching in 3d with mast3r},
  author={Leroy, Vincent and Cabon, Yohann and Revaud, J{\'e}r{\^o}me},
  booktitle={European Conference on Computer Vision},
  pages={71--91},
  year={2024},
  organization={Springer}
}

@inproceedings{kirillov2023segment,
  title={Segment anything},
  author={Kirillov, Alexander and Mintun, Eric and Ravi, Nikhila and Mao, Hanzi and Rolland, Chloe and Gustafson, Laura and Xiao, Tete and Whitehead, Spencer and Berg, Alexander C and Lo, Wan-Yen and others},
  booktitle={Proceedings of the IEEE/CVF international conference on computer vision},
  pages={4015--4026},
  year={2023}
}

@inproceedings{karaev2024cotracker,
  title={Cotracker: It is better to track together},
  author={Karaev, Nikita and Rocco, Ignacio and Graham, Benjamin and Neverova, Natalia and Vedaldi, Andrea and Rupprecht, Christian},
  booktitle={European conference on computer vision},
  pages={18--35},
  year={2024},
  organization={Springer}
}

@article{xu2024das3r,
  title={Das3r: Dynamics-aware gaussian splatting for static scene reconstruction},
  author={Xu, Kai and Tse, Tze Ho Elden and Peng, Jizong and Yao, Angela},
  journal={arXiv preprint arXiv:2412.19584},
  year={2024}
}

@inproceedings{zhang2025monst3r,
  title={MonST3R: A Simple Approach for Estimating Geometry in the Presence of Motion},
  author={Zhang, Junyi and Herrmann, Charles and Hur, Junhwa and Jampani, Varun and Darrell, Trevor and Cole, Forrester and Sun, Deqing and Yang, Ming-Hsuan},
  booktitle={ICLR},
  year={2025}
}

@inproceedings{chen2025easi3r,
  title={Easi3r: Estimating disentangled motion from dust3r without training},
  author={Chen, Xingyu and Chen, Yue and Xiu, Yuliang and Geiger, Andreas and Chen, Anpei},
  booktitle={Proceedings of the IEEE/CVF International Conference on Computer Vision},
  pages={9158--9168},
  year={2025}
}

@article{qi2025strengthening,
  title={Strengthening Generative Robot Policies through Predictive World Modeling},
  author={Qi, Han and Yin, Haocheng and Du, Yilun and Yang, Heng},
  journal={arXiv preprint arXiv:2502.00622},
  year={2025}
}

@inproceedings{wang2025vggt,
  title={Vggt: Visual geometry grounded transformer},
  author={Wang, Jianyuan and Chen, Minghao and Karaev, Nikita and Vedaldi, Andrea and Rupprecht, Christian and Novotny, David},
  booktitle={Proceedings of the Computer Vision and Pattern Recognition Conference},
  pages={5294--5306},
  year={2025}
}

@inproceedings{zhou2026page,
  title={Page-4d: Disentangled pose and geometry estimation for vggt-4d perception},
  author={Zhou, Kaichen and Wang, Yuhan and Chen, Grace and Beaudouin, Gaspard and Zhan, Fangneng and Liang, Paul and Wang, Mengyu},
  booktitle={International Conference on Learning Representations},
  volume={2026},
  pages={36401--36414},
  year={2026}
}

@inproceedings{wang2025continuous,
  title={Continuous 3d perception model with persistent state},
  author={Wang, Qianqian and Zhang, Yifei and Holynski, Aleksander and Efros, Alexei A and Kanazawa, Angjoo},
  booktitle={Proceedings of the Computer Vision and Pattern Recognition Conference},
  pages={10510--10522},
  year={2025}
}

@inproceedings{zhou2026rise,
  title={RISE: Single Static Radar-based Indoor Scene Understanding},
  author={Zhou, Kaichen and Dodds, Laura and Afzal, Sayed Saad and Adib, Fadel},
  booktitle={Proceedings of the IEEE/CVF Conference on Computer Vision and Pattern Recognition},
  pages={32194--32205},
  year={2026}
}

@article{wu2025geometry,
  title={Geometry forcing: Marrying video diffusion and 3d representation for consistent world modeling},
  author={Wu, Haoyu and Wu, Diankun and He, Tianyu and Guo, Junliang and Ye, Yang and Duan, Yueqi and Bian, Jiang},
  journal={arXiv preprint arXiv:2507.07982},
  year={2025}
}

@article{zhang2025videorepa,
  title={VideoREPA: Learning Physics for Video Generation through Relational Alignment with Foundation Models},
  author={Zhang, Xiangdong and Liao, Jiaqi and Zhang, Shaofeng and Meng, Fanqing and Wan, Xiangpeng and Yan, Junchi and Cheng, Yu},
  journal={arXiv preprint arXiv:2505.23656},
  year={2025}
}

@article{zhen2025tesseract,
  title={TesserAct: learning 4D embodied world models},
  author={Zhen, Haoyu and Sun, Qiao and Zhang, Hongxin and Li, Junyan and Zhou, Siyuan and Du, Yilun and Gan, Chuang},
  journal={arXiv preprint arXiv:2504.20995},
  year={2025}
}

@article{wan2025wan,
  title={Wan: Open and advanced large-scale video generative models},
  author={Wan, Team and Wang, Ang and Ai, Baole and Wen, Bin and Mao, Chaojie and Xie, Chen-Wei and Chen, Di and Yu, Feiwu and Zhao, Haiming and Yang, Jianxiao and others},
  journal={arXiv preprint arXiv:2503.20314},
  year={2025}
}

@article{wu2024ivideogpt,
  title={ivideogpt: Interactive videogpts are scalable world models},
  author={Wu, Jialong and Yin, Shaofeng and Feng, Ningya and He, Xu and Li, Dong and Hao, Jianye and Long, Mingsheng},
  journal={Advances in Neural Information Processing Systems},
  volume={37},
  pages={68082--68119},
  year={2024}
}

@article{black2024pi0,
  title={pi0: A Vision-Language-Action Flow Model for General Robot Control},
  author={Black, Kevin and Brown, Noah and Driess, Danny and Esmail, Adnan and Equi, Michael and Finn, Chelsea and Fusai, Niccolo and Groom, Lachy and Hausman, Karol and Ichter, Brian and others},
  journal={arXiv preprint arXiv:2410.24164},
  year={2024}
}

@article{yang2025fast3r,
  title={Fast3R: Towards 3D Reconstruction of 1000+ Images in One Forward Pass},
  author={Yang, Jianing and Sax, Alexander and Liang, Kevin J and Henaff, Mikael and Tang, Hao and Cao, Ang and Chai, Joyce and Meier, Franziska and Feiszli, Matt},
  journal={arXiv preprint arXiv:2501.13928},
  year={2025}
}

@article{kim2024openvla,
  title={Openvla: An open-source vision-language-action model},
  author={Kim, Moo Jin and Pertsch, Karl and Karamcheti, Siddharth and Xiao, Ted and Balakrishna, Ashwin and Nair, Suraj and Rafailov, Rafael and Foster, Ethan and Lam, Grace and Sanketi, Pannag and others},
  journal={arXiv preprint arXiv:2406.09246},
  year={2024}
}

@article{chen2024diffusion,
  title={Diffusion forcing: Next-token prediction meets full-sequence diffusion},
  author={Chen, Boyuan and Mart{\'\i} Mons{\'o}, Diego and Du, Yilun and Simchowitz, Max and Tedrake, Russ and Sitzmann, Vincent},
  journal={Advances in Neural Information Processing Systems},
  volume={37},
  pages={24081--24125},
  year={2024}
}

@article{huang2025self,
  title={Self forcing: Bridging the train-test gap in autoregressive video diffusion},
  author={Huang, Xun and Li, Zhengqi and He, Guande and Zhou, Mingyuan and Shechtman, Eli},
  journal={arXiv preprint arXiv:2506.08009},
  year={2025}
}

@article{yang2024cogvideox,
  title={Cogvideox: Text-to-video diffusion models with an expert transformer},
  author={Yang, Zhuoyi and Teng, Jiayan and Zheng, Wendi and Ding, Ming and Huang, Shiyu and Xu, Jiazheng and Yang, Yuanming and Hong, Wenyi and Zhang, Xiaohan and Feng, Guanyu and others},
  journal={arXiv preprint arXiv:2408.06072},
  year={2024}
}

@article{huang2026pointworld,
  title={PointWorld: Scaling 3D World Models for In-The-Wild Robotic Manipulation},
  author={Huang, Wenlong and Chao, Yu-Wei and Mousavian, Arsalan and Liu, Ming-Yu and Fox, Dieter and Mo, Kaichun and Fei-Fei, Li},
  journal={arXiv preprint arXiv:2601.03782},
  year={2026}
}

@inproceedings{karaev2025cotracker3,
  title={Cotracker3: Simpler and better point tracking by pseudo-labelling real videos},
  author={Karaev, Nikita and Makarov, Yuri and Wang, Jianyuan and Neverova, Natalia and Vedaldi, Andrea and Rupprecht, Christian},
  booktitle={Proceedings of the IEEE/CVF International Conference on Computer Vision},
  pages={6013--6022},
  year={2025}
}

@article{khazatsky2024droid,
  title={Droid: A large-scale in-the-wild robot manipulation dataset},
  author={Khazatsky, Alexander and Pertsch, Karl and Nair, Suraj and Balakrishna, Ashwin and Dasari, Sudeep and Karamcheti, Siddharth and Nasiriany, Soroush and Srirama, Mohan Kumar and Chen, Lawrence Yunliang and Ellis, Kirsty and others},
  journal={arXiv preprint arXiv:2403.12945},
  year={2024}
}

@article{james2020rlbench,
  title={Rlbench: The robot learning benchmark \& learning environment},
  author={James, Stephen and Ma, Zicong and Arrojo, David Rovick and Davison, Andrew J},
  journal={IEEE Robotics and Automation Letters},
  volume={5},
  number={2},
  pages={3019--3026},
  year={2020},
  publisher={IEEE}
}

@article{tao2024maniskill3,
  title={Maniskill3: Gpu parallelized robotics simulation and rendering for generalizable embodied ai},
  author={Tao, Stone and Xiang, Fanbo and Shukla, Arth and Qin, Yuzhe and Hinrichsen, Xander and Yuan, Xiaodi and Bao, Chen and Lin, Xinsong and Liu, Yulin and Chan, Tse-kai and others},
  journal={arXiv preprint arXiv:2410.00425},
  year={2024}
}

@inproceedings{walke2023bridgedata,
  title={Bridgedata v2: A dataset for robot learning at scale},
  author={Walke, Homer Rich and Black, Kevin and Zhao, Tony Z and Vuong, Quan and Zheng, Chongyi and Hansen-Estruch, Philippe and He, Andre Wang and Myers, Vivek and Kim, Moo Jin and Du, Max and others},
  booktitle={Conference on Robot Learning},
  pages={1723--1736},
  year={2023},
  organization={PMLR}
}

@article{brohan2022rt,
  title={Rt-1: Robotics transformer for real-world control at scale},
  author={Brohan, Anthony and Brown, Noah and Carbajal, Justice and Chebotar, Yevgen and Dabis, Joseph and Finn, Chelsea and Gopalakrishnan, Keerthana and Hausman, Karol and Herzog, Alex and Hsu, Jasmine and others},
  journal={arXiv preprint arXiv:2212.06817},
  year={2022}
}

@inproceedings{bruce2024genie,
  title={Genie: Generative interactive environments},
  author={Bruce, Jake and Dennis, Michael D and Edwards, Ashley and Parker-Holder, Jack and Shi, Yuge and Hughes, Edward and Lai, Matthew and Mavalankar, Aditi and Steigerwald, Richie and Apps, Chris and others},
  booktitle={Forty-first International Conference on Machine Learning},
  year={2024}
}

@article{zhi20253dflowaction,
  title={3dflowaction: Learning cross-embodiment manipulation from 3d flow world model},
  author={Zhi, Hongyan and Chen, Peihao and Zhou, Siyuan and Dong, Yubo and Wu, Quanxi and Han, Lei and Tan, Mingkui},
  journal={arXiv preprint arXiv:2506.06199},
  year={2025}
}

@article{liu2025robotransfer,
  title={RoboTransfer: Controllable Geometry-Consistent Video Diffusion for Manipulation Policy Transfer},
  author={Liu, Liu and Wang, Xiaofeng and Zhao, Guosheng and Li, Keyu and Qin, Wenkang and Zhu, Jiagang and Qiu, Jiaxiong and Zhu, Zheng and Huang, Guan and Su, Zhizhong},
  journal={arXiv preprint arXiv:2505.23171},
  year={2025}
}

@article{liu2025geometry,
  title={Geometry-aware 4d video generation for robot manipulation},
  author={Liu, Zeyi and Li, Shuang and Cousineau, Eric and Feng, Siyuan and Burchfiel, Benjamin and Song, Shuran},
  journal={arXiv preprint arXiv:2507.01099},
  year={2025}
}

@article{zhang2025imowm,
  title={imowm: Taming interactive multi-modal world model for robotic manipulation},
  author={Zhang, Chuanrui and Wu, Zhengxian and Lu, Guanxing and Tang, Yansong and Wang, Ziwei},
  journal={arXiv preprint arXiv:2510.09036},
  year={2025}
}

@article{chi2025mind,
  title={MinD: Learning A Dual-System World Model for Real-Time Planning and Implicit Risk Analysis},
  author={Chi, Xiaowei and Ge, Kuangzhi and Liu, Jiaming and Zhou, Siyuan and Jia, Peidong and He, Zichen and Liu, Yuzhen and Li, Tingguang and Han, Lei and Han, Sirui and others},
  journal={arXiv preprint arXiv:2506.18897},
  year={2025}
}

@inproceedings{wang2024dust3r,
  title={Dust3r: Geometric 3d vision made easy},
  author={Wang, Shuzhe and Leroy, Vincent and Cabon, Yohann and Chidlovskii, Boris and Revaud, Jerome},
  booktitle={Proceedings of the IEEE/CVF conference on computer vision and pattern recognition},
  pages={20697--20709},
  year={2024}
}

@article{guo2025ctrl,
  title={Ctrl-world: A controllable generative world model for robot manipulation},
  author={Guo, Yanjiang and Shi, Lucy Xiaoyang and Chen, Jianyu and Finn, Chelsea},
  journal={arXiv preprint arXiv:2510.10125},
  year={2025}
}

@article{liu2024sora,
  title={Sora: A review on background, technology, limitations, and opportunities of large vision models},
  author={Liu, Yixin and Zhang, Kai and Li, Yuan and Yan, Zhiling and Gao, Chujie and Chen, Ruoxi and Yuan, Zhengqing and Huang, Yue and Sun, Hanchi and Gao, Jianfeng and others},
  journal={arXiv preprint arXiv:2402.17177},
  year={2024}
}

@article{lin2025depth,
  title={Depth anything 3: Recovering the visual space from any views},
  author={Lin, Haotong and Chen, Sili and Liew, Junhao and Chen, Donny Y and Li, Zhenyu and Shi, Guang and Feng, Jiashi and Kang, Bingyi},
  journal={arXiv preprint arXiv:2511.10647},
  year={2025}
}

@article{lipman2022flow,
  title={Flow matching for generative modeling},
  author={Lipman, Yaron and Chen, Ricky TQ and Ben-Hamu, Heli and Nickel, Maximilian and Le, Matt},
  journal={arXiv preprint arXiv:2210.02747},
  year={2022}
}

@article{kingma2013auto,
  title={Auto-encoding variational bayes},
  author={Kingma, Diederik P and Welling, Max},
  journal={arXiv preprint arXiv:1312.6114},
  year={2013}
}

@article{wu2025qwen,
  title={Qwen-image technical report},
  author={Wu, Chenfei and Li, Jiahao and Zhou, Jingren and Lin, Junyang and Gao, Kaiyuan and Yan, Kun and Yin, Sheng-ming and Bai, Shuai and Xu, Xiao and Chen, Yilei and others},
  journal={arXiv preprint arXiv:2508.02324},
  year={2025}
}

@article{ravi2024sam,
  title={Sam 2: Segment anything in images and videos},
  author={Ravi, Nikhila and Gabeur, Valentin and Hu, Yuan-Ting and Hu, Ronghang and Ryali, Chaitanya and Ma, Tengyu and Khedr, Haitham and R{\"a}dle, Roman and Rolland, Chloe and Gustafson, Laura and others},
  journal={arXiv preprint arXiv:2408.00714},
  year={2024}
}

@inproceedings{wen2024foundationpose,
  title={Foundationpose: Unified 6d pose estimation and tracking of novel objects},
  author={Wen, Bowen and Yang, Wei and Kautz, Jan and Birchfield, Stan},
  booktitle={Proceedings of the IEEE/CVF conference on computer vision and pattern recognition},
  pages={17868--17879},
  year={2024}
}

@article{unterthiner2019fvd,
  title={FVD: A new metric for video generation},
  author={Unterthiner, Thomas and Van Steenkiste, Sjoerd and Kurach, Karol and Marinier, Rapha{\"e}l and Michalski, Marcin and Gelly, Sylvain},
  year={2019}
}

@article{wang2004image,
  title={Image quality assessment: from error visibility to structural similarity},
  author={Wang, Zhou and Bovik, Alan C and Sheikh, Hamid R and Simoncelli, Eero P},
  journal={IEEE transactions on image processing},
  volume={13},
  number={4},
  pages={600--612},
  year={2004},
  publisher={IEEE}
}

@article{zhou2023dynpoint,
  title={Dynpoint: Dynamic neural point for view synthesis},
  author={Zhou, Kaichen and Zhong, Jia-Xing and Shin, Sangyun and Lu, Kai and Yang, Yiyuan and Markham, Andrew and Trigoni, Niki},
  journal={Advances in Neural Information Processing Systems},
  volume={36},
  pages={69532--69545},
  year={2023}
}

@inproceedings{peebles2023scalable,
  title={Scalable diffusion models with transformers},
  author={Peebles, William and Xie, Saining},
  booktitle={Proceedings of the IEEE/CVF international conference on computer vision},
  pages={4195--4205},
  year={2023}
}

@article{chen2025large,
  title={Large video planner enables generalizable robot control},
  author={Chen, Boyuan and Zhang, Tianyuan and Geng, Haoran and Song, Kiwhan and Zhang, Caiyi and Li, Peihao and Freeman, William T and Malik, Jitendra and Abbeel, Pieter and Tedrake, Russ and others},
  journal={arXiv preprint arXiv:2512.15840},
  year={2025}
}

@article{fu2025learning,
  title={Learning video generation for robotic manipulation with collaborative trajectory control},
  author={Fu, Xiao and Wang, Xintao and Liu, Xian and Bai, Jianhong and Xu, Runsen and Wan, Pengfei and Zhang, Di and Lin, Dahua},
  journal={arXiv preprint arXiv:2506.01943},
  year={2025}
}

@article{qian2025wristworld,
  title={Wristworld: Generating wrist-views via 4d world models for robotic manipulation},
  author={Qian, Zezhong and Chi, Xiaowei and Li, Yuming and Wang, Shizun and Qin, Zhiyuan and Ju, Xiaozhu and Han, Sirui and Zhang, Shanghang},
  journal={arXiv preprint arXiv:2510.07313},
  year={2025}
}

@article{huang2026skill,
  title={Skill-Aware Diffusion for Generalizable Robotic Manipulation},
  author={Huang, Aoshen and Chen, Jiaming and Cheng, Jiyu and Song, Ran and Pan, Wei and Zhang, Wei},
  journal={arXiv preprint arXiv:2601.11266},
  year={2026}
}

@article{wu2026multiworld,
  title={MultiWorld: Scalable Multi-Agent Multi-View Video World Models},
  author={Wu, Haoyu and Yu, Jiwen and Zou, Yingtian and Liu, Xihui},
  journal={arXiv preprint arXiv:2604.18564},
  year={2026}
}

@article{li2025latent,
  title={4D Latent World Model for Robot Planning},
  author={Li, Zhiyi and Wu, Peilin and Han, Xiaoshen and Cai, Ruojin and Du, Yilun},
  year={2026}
}

@article{feng2025vidarc,
  title={Vidarc: Embodied Video Diffusion Model for Closed-loop Control},
  author={Feng, Yao and Xiang, Chendong and Mao, Xinyi and Tan, Hengkai and Zhang, Zuyue and Huang, Shuhe and Zheng, Kaiwen and Liu, Haitian and Su, Hang and Zhu, Jun},
  journal={arXiv preprint arXiv:2512.17661},
  year={2025}
}

@inproceedings{yu2025repa,
  title     = {Representation Alignment for Generation: Training Diffusion Transformers Is Easier Than You Think},
  author    = {Yu, Sihyun and Kwak, Sangkyung and Jang, Huiwon and Jeong, Jongheon and Huang, Jonathan and Shin, Jinwoo and Xie, Saining},
  booktitle = {International Conference on Learning Representations (ICLR)},
  year      = {2025}
}

@inproceedings{bharadhwaj2024track2act,
  title     = {Track2Act: Predicting Point Tracks from Internet Videos Enables Generalizable Robot Manipulation},
  author    = {Bharadhwaj, Homanga and Mottaghi, Roozbeh and Gupta, Abhinav and Tulsiani, Shubham},
  booktitle = {European Conference on Computer Vision (ECCV)},
  year      = {2024}
}

@article{blattmann2023stable,
  title   = {Stable Video Diffusion: Scaling Latent Video Diffusion Models to Large Datasets},
  author  = {Blattmann, Andreas and Dockhorn, Tim and Kulal, Sumith and Mendelevitch, Daniel and Kilian, Maciej and Lorenz, Dominik and Levi, Yam and English, Zion and Voleti, Vikram and Letts, Adam and others},
  journal = {arXiv preprint arXiv:2311.15127},
  year    = {2023}
}

@inproceedings{wen2024atm,
  title     = {Any-point Trajectory Modeling for Policy Learning},
  author    = {Wen, Chuan and Lin, Xingyu and So, John and Chen, Kai and Dou, Qi and Gao, Yang and Abbeel, Pieter},
  booktitle = {Robotics: Science and Systems (RSS)},
  year      = {2024}
}

@article{bjorck2025gr00t,
  title   = {{GR00T N1}: An Open Foundation Model for Generalist Humanoid Robots},
  author  = {Bj{\"o}rck, Johan and Casta{\~n}eda, Fernando and Cherniadev, Nikita and Da, Xingye and Ding, Runyu and Fan, Linxi and Fang, Yu and Fox, Dieter and Hu, Fengyuan and Huang, Spencer and others},
  journal = {arXiv preprint arXiv:2503.14734},
  year    = {2025}
}

@inproceedings{wang2024spann3r,
  title     = {3D Reconstruction with Spatial Memory},
  author    = {Wang, Hengyi and Agapito, Lourdes},
  booktitle = {International Conference on 3D Vision (3DV)},
  year      = {2025}
}

@article{wang2025pi3,
  title   = {{$\pi^3$}: Scalable Permutation-Equivariant Visual Geometry Learning},
  author  = {Wang, Yifan and Zhou, Jianjun and Zhu, Haoyi and Chang, Wenzheng and Zhou, Yang and Li, Zizun and Chen, Junyi and Pang, Jiangmiao and Shen, Chunhua and He, Tong},
  journal = {arXiv preprint arXiv:2507.13347},
  year    = {2025}
}

@article{murali2025graspgen,
  title   = {{GraspGen}: A Diffusion-Based Framework for 6-{DOF} Grasping with On-Generator Training},
  author  = {Murali, Adithyavairavan and Sundaralingam, Balakumar and Chao, Yu-Wei and Yamada, Jun and Yuan, Wentao and Carlson, Mark and Ramos, Fabio and Birchfield, Stan and Fox, Dieter and Eppner, Clemens},
  journal = {arXiv preprint arXiv:2507.13097},
  year    = {2025}
}

@inproceedings{shoemake1985animating,
  title     = {Animating Rotation with Quaternion Curves},
  author    = {Shoemake, Ken},
  booktitle = {Proceedings of the 12th Annual Conference on Computer Graphics and Interactive Techniques (SIGGRAPH)},
  pages     = {245--254},
  year      = {1985},
  publisher = {ACM}
}

@inproceedings{eigen2014depth,
  title     = {Depth Map Prediction from a Single Image Using a Multi-Scale Deep Network},
  author    = {Eigen, David and Puhrsch, Christian and Fergus, Rob},
  booktitle = {Advances in Neural Information Processing Systems (NeurIPS)},
  year      = {2014}
}

@inproceedings{fan2017point,
  title     = {A Point Set Generation Network for 3D Object Reconstruction from a Single Image},
  author    = {Fan, Haoqiang and Su, Hao and Guibas, Leonidas J.},
  booktitle = {IEEE Conference on Computer Vision and Pattern Recognition (CVPR)},
  pages     = {605--613},
  year      = {2017}
}

@inproceedings{doersch2022tapvid,
  title     = {{TAP-Vid}: A Benchmark for Tracking Any Point in a Video},
  author    = {Doersch, Carl and Gupta, Ankush and Markeeva, Larisa and Recasens, Adri{\`a} and Smaira, Lucas and Aytar, Yusuf and Carreira, Jo{\~a}o and Zisserman, Andrew and Yang, Yi},
  booktitle = {Advances in Neural Information Processing Systems (NeurIPS), Datasets and Benchmarks Track},
  year      = {2022}
}

@article{wang2026vggt,
  title={VGGT-omega},
  author={Wang, Jianyuan and Chen, Minghao and Zhang, Shangzhan and Karaev, Nikita and Sch{\"o}nberger, Johannes and Labatut, Patrick and Bojanowski, Piotr and Novotny, David and Vedaldi, Andrea and Rupprecht, Christian},
  journal={arXiv preprint arXiv:2605.15195},
  year={2026}
}
